\documentclass[a4paper,fleqn]{cas-sc}

\usepackage{amssymb}
\usepackage{relsize}
\usepackage{array}
\usepackage{caption}
\usepackage{subcaption}
\usepackage{amsmath}

\usepackage{tikz}
\usepackage{adjustbox}
\usepackage{amsfonts} 
\usepackage{siunitx}
\usepackage{float}
\usepackage{booktabs}						
\usepackage{multirow}
\usepackage{dsfont}
\usepackage{algorithm}
\usepackage{algpseudocode}

\usepackage{hyperref}

\usepackage{graphicx}
\usepackage{caption}
\usepackage{subcaption}
\usepackage{amssymb}
\usepackage{relsize}
\usepackage{array}
\usepackage{tikz}
\usepackage{adjustbox}
\usepackage{multicol}
\usepackage{xcolor} 
\usepackage{graphicx}
\usepackage{colortbl}
\usepackage{hyperref}
\usepackage{float}

\usepackage[authoryear,longnamesfirst]{natbib}

\def\tsc#1{\csdef{#1}{\textsc{\lowercase{#1}}\xspace}}
\tsc{WGM}
\tsc{QE}

\begin{document}
\let\WriteBookmarks\relax
\def\floatpagepagefraction{1}
\def\textpagefraction{.001}

\shorttitle{A Transformer approach for Electricity Price Forecasting}    

\shortauthors{O.Llorente et~al.}

\title [mode = title]{A Transformer approach for Electricity Price Forecasting} 

\author[inst1]{Oscar Llorente}
\ead{oscar.llorente.gonzalez@ericsson.com}

\affiliation[inst1]{organization={Ericsson Cognitive Labs},
            addressline={Retama Ed 1 Torre Suecia}, 
            city={Madrid},
            postcode={28045}, 
            state={Madrid},
            country={Spain}}

\author[inst2]{Jose Portela}
\ead{jose.portela@iit.comillas.edu}

\affiliation[inst2]{organization={Institute for Research in Technology~(IIT), ICAI School of Engineering, Comillas Pontifical University},
            addressline={Santa Cruz de Marcenado, 26}, 
            city={Madrid},
            postcode={28015}, 
            state={Madrid},
            country={Spain}}

\begin{abstract}
This paper presents a novel approach to electricity price forecasting (EPF) using a pure Transformer model. As opposed to other alternatives, no other recurrent network is used in combination to the attention mechanism. Hence, showing that the attention layer is enough for capturing the temporal patterns. The paper also provides fair comparison of the models using the open-source EPF toolbox and provide the code to enhance reproducibility and transparency in EPF research. The results show that the Transformer model outperforms traditional methods, offering a promising solution for reliable and sustainable power system operation~\footnote{All code to replicate our findings will be available here: \url{https://github.com/osllogon/epf-transformers}}.
\end{abstract}



\maketitle

\section{Introduction}

Forecasting electricity prices plays a pivotal role in the modern power systems landscape. Accurate and efficient price forecasting methods have become increasingly crucial as the electricity market continues to evolve towards a more competitive and deregulated structure ~\cite{weron2014electricity, lagoForecastingDayaheadElectricity2021} . These forecasts serve as a fundamental decision-making tool for various market participants, including power producers, consumers, traders, and grid operators.


Furthermore, with the growing integration of renewable energy sources, which are inherently intermittent and unpredictable, the volatility and complexity of electricity prices have increased. This motivates the need for more sophisticated and robust forecasting methods to capture these complex dynamics and provide accurate price forecasts ~\cite{Bhatia2021ensemble}. Thus, developing effective electricity price forecasting methods is not only an economic imperative but also a key enabler for the sustainable and reliable operation of power systems.

The techniques for electricity price forecasting (EPF) have evolved significantly over the years, with recent advancements in machine learning and artificial intelligence leading the way ~\cite{Zhang202Deep}. While effective in certain scenarios, traditional statistical methods often struggle to capture the complex, non-linear dynamics of electricity prices. This has led to the exploration of more sophisticated models, particularly neural networks and deep learning models applied to time series forecasting. For example, Recurrent Neural Networks (RNNs) and Long Short Term Memory (LSTM) networks have been successfully employed for EPF ~\cite{memarzadeh2021short}. These models are capable of learning complex patterns and long-term dependencies, making them well-suited for forecasting tasks.

Among the deep learning models, the application of Transformer-based architecture is an emerging technique. The design of attention mechanisms for approaching sequence-to-sequence problems started in the NLP field, where the output is not only a label, but a multiple sequence (i.e. a complete sentence). Traditionally an encoder-decoder structure was used until the Attention layer was presented 
\cite{bahdanauNeuralMachineTranslation2016}. The Attention layer allows the decoder to focus its attention on a specific word or group of words from the input of the encoder (the original sentence in the case of translation). This helped improve the State of the Art of many NLP problems, becoming the basis of the Transformer model. One of the advantages of the Transformer architecture is that it enables the generation of much bigger Deep Learning models (such as the BERT model~\cite{Devlin2019BERTPO}).

The attention mechanisms that allow focus on the most relevant parts of the input make them powerful tools for prediction tasks and have been successfully used in several fields such as Autonomous Vehicles~\cite{autonomousdriving} or Image Classification~\cite{dosovitskiyImageWorth16x162022}.

Attention mechanisms have also been applied to the power sector. For example, ~\cite{marulanda2023hybrid} use attention mechanisms combined with LSTM models for short-term wind power forecasting. Other examples include  \cite{zhao2021short} and \cite{chenShortTermLoadForecasting2019} which propose the use of attention and transformers for electricity load forecasting. 

Regarding electricity price forecasting, additional challenges are presented as input explanatory variables have to be included as relevant predictors in the model, hence, 
\cite{bottieauInterpretableTransformerModel2023} uses an attention layer combined with BLSTM layers for electricity price prediction to achieve promising results. Additionally, \cite{liDenseSkipAttention2022} propose a Transformer-based model combined with GRU layers for probabilistic electricity price forecasting.

Therefore, the application of Transformer models in forecasting is still in its early stages, and further research is needed to understand their potential and limitations with respect to other forecasting methods. In fact, most of the mentioned models were not Transformers, but a recurrent model combined with attention layers. Authors such as \cite{zengAreTransformersEffective2023}, still question the effectiveness of Transformer capabilities, hence, making this a relevant issue in the research community.

Additionally, in the EPF field, the issue of reproducibility has been a significant challenge. Many studies have utilized unique, non-public datasets and have tested their methods over too short and limited market samples, making it difficult to evaluate the effectiveness of new predictive algorithms. In response to these issues, ~\cite{lagoForecastingDayaheadElectricity2021} proposed the EPF toolbox, an open-source toolbox for electricity price forecasting that enhances reproducibility, promotes transparency and drives innovation.

As opposed to other approaches, where attention layers are combined with recurrent neural networks, this paper explores the application of Transformers to electricity price forecasting by using a pure Transformer model, showing that the attention layer is enough for capturing the temporal patterns. In addition, a fair benchmark using the EPF toolbox is provided. Moreover, by making the proposed model open-source, we aim to address the reproducibility issue and promote transparency and collaboration in the field of electricity price forecasting.

This paper is organized as follows: 
Section II describes the proposed transformer model for EPF. Section III describes in detail the case study setup and section IV analyzes the results. Finally, section V provides conclusions of the research.

\section{Proposed Transformer model for Electricity Price Forecasting}
\label{sec: proposed model}
This section explains the proposed model for EPF. The objective is the same as stated in \cite{lagoForecastingDayaheadElectricity2021}: to predict the electricity prices of the 24h of the day $D + 1$ based on the exogenous variables of the 24h of the day $D + 1$ as well as past prices and exogenous variables. 

The model proposed in this paper is a pure Transformer, specifically, a Transformer Encoder as in the case of BERT~\cite{devlinBERTPretrainingDeep2019} or ViT~\cite{dosovitskiyImageWorth16x162022}. With this structure, the Transformer can capture temporal patterns and make a prediction focusing more attention on longer trends than other temporal models (e.g. LSTMs). This is evidenced in the case of Natural Language Processing ~\cite{vaswaniAttentionAllYou2017}, where the Transformer does not forget the previous information in longer sequences. 

\begin{figure}
    \centering
    \includegraphics[width=0.3\linewidth, height=0.5\linewidth]{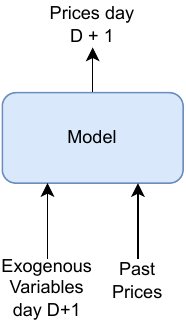}
    \caption{Proposed modeling structure}
    \label{fig: model structure}
\end{figure}

We ought to establish a direct relationship between exogenous variables and prices and keep the Transformer to model the price dynamics. Therefore, it is decided to include only past prices as historical information. Therefore, this model will have two inputs, the exogenous values for the 24h of the day $D + 1$ and the prices of the 24h for the past $m$ days. The general structure can be visualized in Figure~\ref{fig: model structure}. Then, the interval used for the past prices will be [$D - m$, $D$], including both days. These $m$ days will be a hyperparameter of the model called \textit{sequence length} that should be optimized. 

\begin{figure}
  \centering
  \begin{subfigure}{0.49\linewidth}
      \centering
      \includegraphics[width=\linewidth, height=1.8\linewidth]{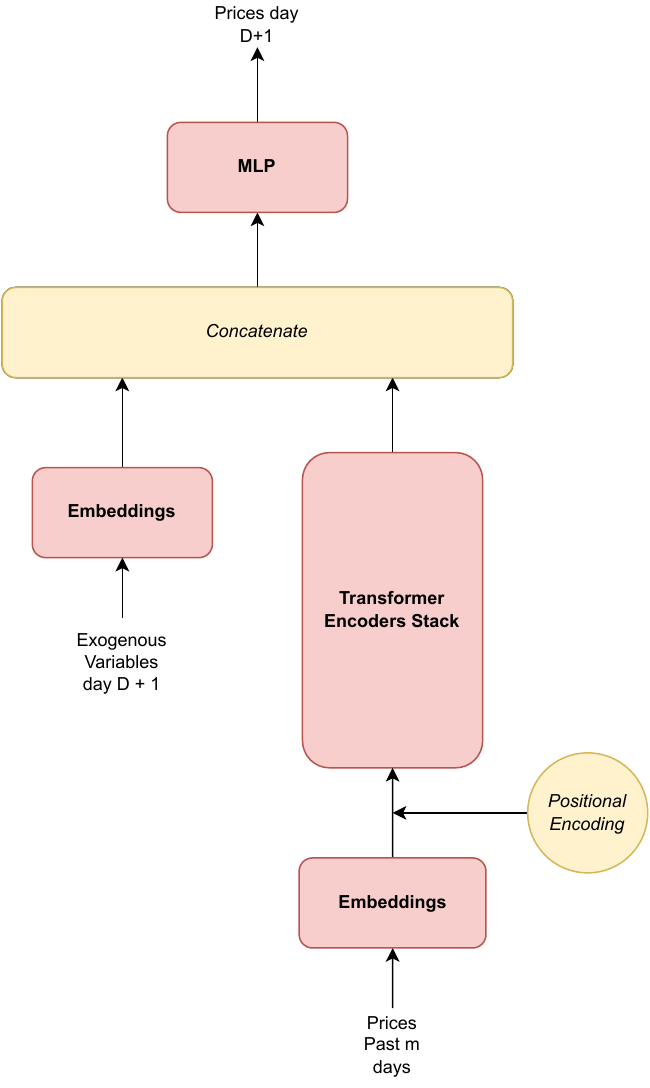}
      \caption{Model Architecture}
      \label{fig: model architecture}
  \end{subfigure}
  \begin{subfigure}{0.49\linewidth}
    \centering
    \includegraphics[width=\linewidth, height=1.8\linewidth]{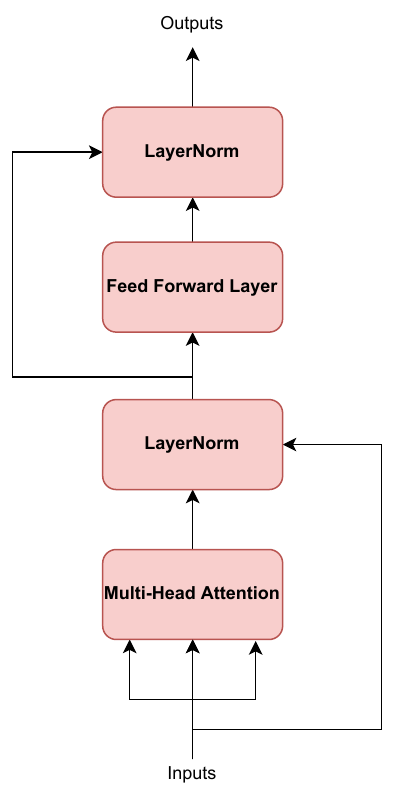}
    \caption{Encoder Architecture}
    \label{fig: encoder architecture}
  \end{subfigure}
  \caption{Model and Transformer Encoder Architectures}
\end{figure}

The main model architecture is illustrated in ~\ref{fig: model architecture}, which consists of two main data flows, one for each input. Then, the results of each flow are concatenated and served as an input to a multivariate Multilayer Perceptron for predicting day ahead prices. The architecture is described ahead in detail.

On the one hand, the processing flow for the exogenous variables of the day $D + 1$ is done through an Embedding layer. On the other hand, the $m x 24$ past prices are also introduced firstly to an Embedding layer. Then, the output is combined with a Positional Encoding and the result is introduced into a stack of Transformer Encoders to account for the temporal dynamics.

Transformer Encoder structure can also be observed in Figure~\ref{fig: encoder architecture}. The inputs of the first Encoder of the stack have dimensions $m$ x $n$ (being $n$ the hyperparameter accounting for the output dimension of the embedding), and the rest of the Encoders have $m$ x $h$, due to the Feed-Forward layer. The dimensions of the outputs of the Transformer stack are $m$ x $h$, being $h$ the hidden dimension for the Feed-Forward layer. Finally, only the last element of this output sequence of $m$ days will be selected for making the final prediction combined with the outputs of the exogenous variables flow. 

As mentioned, the two outputs of the different flows are concatenated and passed into a Muti-Layer Perceptron, having an input dimension of $h + n$. The output of the Multi-Layer Perceptron will be the final prediction of the prices of the 24h of the day $D + 1$ (i.e. a 24-dimensional output layer).

\begin{figure}
  \centering
  \begin{subfigure}{0.49\linewidth}
      \centering
      \includegraphics[width=0.5\linewidth, height=1\linewidth]{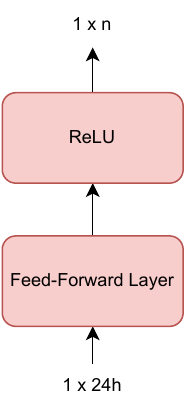}
      \caption{Exogenous Variables}
  \end{subfigure}
  \begin{subfigure}{0.49\linewidth}
    \centering
    \includegraphics[width=0.5\linewidth, height=1\linewidth]{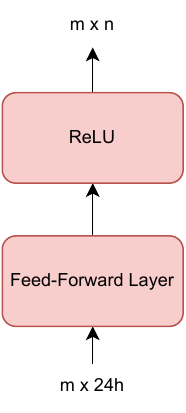}
    \caption{Past prices}
  \end{subfigure}
  \caption{Embeddings}
  \label{fig: embeddings}
\end{figure}

The two Embedding layers defined for preprocessing the inputs are projections into a higher dimensional space, consisting of a Feedforward layer followed by a ReLU. These layers transform each 24-valued vector of one day into a higher-order vector. This is illustrated in Figure~\ref{fig: embeddings}. 

Consequently, the model has four main structures, two types of Embeddings, a stack of Transformer Encoders and a final Multi-Layer Perceptron for making the final predictions. 
In order to train the proposed model architecture, the following hyperparameters will be tuned to optimize the model performance in a validation set:

\begin{itemize}
    \item $n$: Embedding dimension. This element is the size of the vectors the prices will be converted to. 
    \item NH: Number of heads of the attention layers of the Transformer Architecture.
    \item NL: Number of Transformer encoder blocks (shown in Figure \ref{fig: encoder architecture}).
    \item FD: Feed-forward dimension. Dimension of feed-forward layer in the Transformer block.
    \item P: Dropout probability.
    \item Opt: Optimizer used for the optimization. Adam was used. 
    \item SL: Historical Sequence length used for the prediction, i.e., number of past prices per hour used for the prediction. Since only complete days are used every number for this parameter has to be the number of past days $m$ times 24.
    \item Lr: Learning rate used for the optimization.
    \item Sch: Scheduler used for the optimization. Only two values have been used, a step learning rate scheduler, where the first number given is the step and the second the gamma parameter, or None meaning that no scheduler was used.
    \item CG: Clip gradients: Threshold to clip gradients during optimization.
\end{itemize}

\section{Case Study}
\label{sec: case study}
As pointed out in~\cite{lagoForecastingDayaheadElectricity2021}, in the literature the comparisons between different types of models have been unfair most times. They detected the following problems in the published literature on EPF models: 

\begin{itemize}
  \item The datasets in most cases are not the same, and therefore the comparison with other models is unfeasible, having in most cases only comparisons against base models and not with more recent publications.
  \item Test datasets were not long enough. If these datasets are shorter than a year, the behavior of the model could be biased and be better only for a specific time of the year, e.g. a specific month. 
  \item No statistical testing was performed, as the Diebold-Mariano test, to check if the improvements in predictions are significantly better than with other models.
\end{itemize}

Hence they proposed a new framework for testing the different models: epftoolbox~\cite{WelcomeEpftoolboxDocumentation}. This python library allows to use the models they have developed and also test the different datasets to build a fair comparison, solving all the issues that have been mentioned previously. Therefore, in this paper, this framework is used to compare and test the model presented.

\subsection{Open Access Datasets}

To find the best algorithm for electricity price forecasting, it is important to consider multiple datasets, rather than relying on a single one. In \cite{lagoForecastingDayaheadElectricity2021}, five different datasets for different countries are available so methods can be compared fairly: 
\begin{itemize}
    \item Nord Pool (NP), from the European Power Market of the Nordic Countries.
    \item PJM, obtained from Pennsylvania-New Jersey-Maryland market in the United States.
    \item EPEX-BE, obtained from the day-ahead Electricity Market in Belgium.
    \item EPEX-FR, obtained from the day-ahead Electricity Market in France.
    \item EPEX-DE, obtained from the German Electricity Market. 
\end{itemize}
Each dataset contains historical prices plus two time series representing exogenous inputs.

\subsection{Train-Validation-Test split}

The testing period is a source of discussion. Most articles try to evaluate algorithms in short periods, e.g. a week or a month. However, the dynamics in a market can change throughout the year, especially in the electricity domain, which is highly affected by the time of the year due to issues such as the available amount of renewable energies or the time consumers are spending at home (great difference between summer and winter). 
Following \cite{lagoForecastingDayaheadElectricity2021}, a two-year testing period is used to account the different dynamics that can appear.


In particular, these will be the splits between training (train and validation) and test sets:
\begin{table}[h!]
    \begin{center}
        \begin{tabular}{| l | c | c |}
            \textbf{Market} & \textbf{Train period} & \textbf{Test period} \\ \hline
            Nord Pool & 01.01.2013 - 26.12.2016 & 27.12.2016 - 24.12.2018 \\
            PJM & 01.01.2013 - 26.12.2016 & 27.12.2016 - 24.12.2018 \\
            EPEX-BE & 09.01.2011 - 03.01.2015 & 04.01.2015 - 31.12.2016 \\
            EPEX-FR & 09.01.2011 - 03.01.2015 & 04.01.2015 - 31.12.2016 \\ 
            EXPEX-DE & 09.01.2012 - 03.01.2016 & 04.01.2016 - 31.12.2017
        \end{tabular}
        \caption{Train-Test split}
        \label{tab: train-test split}
    \end{center}
\end{table}

\subsection{Retraining and Renormalization}
\label{sec: case study, retraining and renormalization}
In \cite{lagoForecastingDayaheadElectricity2021}, the authors state that a model should be retrained instead of being directly evaluated in the whole dataset, or otherwise it is not tested in realistic conditions, since in real-life forecasting models are often retrained to account for the last market information. Even though this idea is logical, it generates two main problems for testing more advanced models such as the one proposed in this paper

On the one hand, although retraining each day simulate well real life behaviour, there are some difficulties to automatize that retraining for more complicated models. While retraining a Deep Neural Network composed of linear layers and activation functions seems to follow a straightforward process explained in \cite{lagoForecastingDayaheadElectricity2021}, where early-stopping gives the optimal training point for the selected hyperparameters, more complicated models as Transformers can be more difficult to train due to issues like vanishing or exploding gradients or the amplification effect pointed put in \cite{liuUnderstandingDifficultyTraining2020}. Therefore for these models, it is more likely that a Machine Learning Engineer monitors the retraining phase and execute it if necessary. It could be the case that for some time the parameters for the retraining are the same, and only when the data changes are significant enough it is necessary to look again for optimal values.




On the other hand, the computational burden is expensive for advanced models. While the best model in \cite{lagoForecastingDayaheadElectricity2021} takes around five minutes to train, the Transformer model presented in this paper takes around twenty minutes (measured in a computer with a GPU RTX 3060). 


As a consequence of the aforementioned issues, it was decided to evaluate the fitted models in the test periods without retraining. 

Nevertheless, as suggested in \cite{lagoForecastingDayaheadElectricity2021}, a renormalization step was applied consisting on subtracting the mean and standard deviation from the data previous to the next day to predict. This helps the model to adapt if the conditions vary greatly during the test period.

\subsection{Models to Compare}
\label{sec: case study, models to compare}
For the case study presented, three EPF models are tested. evaluating the results from the model presented in this paper two benchmarks will be used:
\begin{itemize}
    \item \textbf{DNN:} Regarding neural networks in ~\cite{lagoForecastingDayaheadElectricity2021}, a Deep Neural Network with two layers was selected as the best Machine Learning Model. In fact, the model that proved to be more accurate in their study was the Ensemble of Deep Neural Networks. Therefore, this model ensemble is selected as a benchmark. As it was explained the ensemble will be trained on the training period and evaluated during the test period using a renormalization step but not retraining.  We compute two approaches, a renormalization with all the past data and a renormalization with only the last year values (as in  \cite{lagoForecastingDayaheadElectricity2021}). 
    \item \textbf{Naïve:}  Because the models are being tested without the retraining step, a naïve model is also selected as a benchmark. This test checks if predictions remain accurate with a substantial time gap between training and prediction.
    \item \textbf{Transformer model:} This is the transformer model proposed. As mentioned in Section~\ref{sec: proposed model}, some hyperparameters have to be optimized using a validation set. Following what was proposed in~\cite{lagoForecastingDayaheadElectricity2021}, validation is constructed with the last 42 weeks of the training set. Some values were proposed for the different hyperparameters and datasets and the configuration with the best performance is selected. In Tables \ref{tab: nord pool validation set results}, \ref{tab: pjm validation set results}, \ref{tab: epex-fr validation set results}, \ref{tab: epex-de validation set results} and \ref{tab: epex-be validation set results} the results of some of the configurations in the different datasets are presented along with the best ones highlighted in gray. The best model is evaluated then on the test set, without any retraining but with data renormalization with all past values.
\end{itemize}

Following what was exposed in \cite{lagoForecastingDayaheadElectricity2021}, to measure the results offered by the model presented in this paper and compare it with the benchmarks explained, the following metrics will be used: MAE, RMSE and sMAPE. 

\subsection{Statistical Testing}
\label{sec: case study, definition, statical testing}
In order to evaluate the statistical significance of the differences in forecasting accuracy between two models,  \textit{Diebold-Mariano test} (DM test) is commonly applied. 

In particular, the one-sided DM test is implemented in~\cite{lagoForecastingDayaheadElectricity2021} and used in the case study.  The null hypothesis of the test is that the predictions of model A are more accurate than the ones of a model B. The lower the p-value obtained, the more observed data inconsistent with the null hypothesis. Taking into account the convention to reject the null hypothesis at p-values lower than 5\%, if a lower p-value is obtained, it can be declared that the difference between predictions of model B and predictions of model A are statistically significant, being B the model with better performance.

\subsection{Software}
\label{sec: case study, definition, software}
The aim of this paper is to compare the Transformer model presented with the ones developed in the ~\cite{lagoForecastingDayaheadElectricity2021}. Hence,  every implementation from the mentioned library will be used in order to follow a rigorous approach to compare the different methods.
As the aim of this paper is to openly contribute to EPF research, all the code and the best models can be found in \url{https://github.com/osllogon/epf_transformers}.

\section{Results}
\label{sec: results}
The results of the selected models tested on the different datasets are presented in this section. For the comparison, the different metrics explained in the Section~\ref{sec: case study, models to compare} are calculated for the test set predictions. 

Tables \ref{tab: naive model results}, \ref{tab: dnn ensemble renormalized with all past values results} and \ref{tab: dnn ensemble renormalized with last-year values results} show the benchmark Naïve, DNN with all past renormalization and DNN with last-year renormalization model results, respectively. As can be observed, in some cases the results of the DNN model are better and, in other cases, the naive model outperforms the DNN Ensembles. This could be due to the fact that the DNN model has to make predictions for a long test period without retraining.

The results of the proposed Transformer model are presented in Table~\ref{tab: final results}. In addition, the DM tests comparing the Transformer against the other benchmark models are also shown. Transformer model has significantly better results (DM test under 0.05) than the model from \cite{lagoForecastingDayaheadElectricity2021} in four of the five datasets. It also has significantly better results than the naive model in all the datasets. 
The results evidence the improvement in adding the transformer architecture to the price time series.

\begin{table}
    \centering
    \caption{Naive model test results}
    \label{tab: naive model results}
    \begin{tabular}{cccc}
        \toprule
          & MAE & RMSE & sMAPE \\
          \midrule
          Nord Pool & 2.89 & 5.31 & 0.08 \\ 
          PJM & 4.43 & 7.24 & 0.17 \\ 
          EPEX-BE & 9.38 & 20.97 & 0.23 \\ 
          EPEX-FR & 7.24 & 15.2 & 0.22 \\ 
          EPEX-DE & 8.19 & 13.19 & 0.3 \\
        \bottomrule
    \end{tabular}
\end{table}

\begin{table}[h]
    \centering
    \caption{DNN Ensemble renormalized with all past values test results}
    \label{tab: dnn ensemble renormalized with all past values results}
    \begin{tabular}{cccc}
        \toprule
          & MAE & RMSE & sMAPE \\
          \midrule
          Nord Pool & 3.4 & 5.27 & 0.09 \\ 
          PJM & 7.17 & 10.24 & 0.27 \\
          EPEX-BE & 9.38 & 18.49 & 0.24 \\ 
          EPEX-FR & 4.74 & 12.11 & 0.14 \\ 
          EPEX-DE & 4.34 & 7.63 & 0.17 \\
        \bottomrule
    \end{tabular}
\end{table}

\begin{table}
    \centering
    \caption{DNN Ensemble renormalized with last-year values test results}
    \label{tab: dnn ensemble renormalized with last-year values results}
    \begin{tabular}{cccc}
        \toprule
          & MAE & RMSE & sMAPE \\
          \midrule
          Nord Pool & 3.78 & 5.73 & 0.09 \\ 
          PJM & 11.98 & 17.92 & 0.35 \\ 
          EPEX-BE & 10.87 & 20.68 & 0.28 \\ 
          EPEX-FR & 4.58 & 11.99 & 0.13 \\ 
          EPEX-DE & 4.38 & 8.07 & 0.17 \\
        \bottomrule
    \end{tabular}
\end{table}

\begin{table}[h]
    \centering
    \caption{Transformer model test results and DM p-values obtained.}
    \label{tab: final results}
     \scalebox{0.55}{
        \begin{tabular}{ccccccc}
            \toprule
              \textbf{Market} & \textbf{MAE} & \textbf{RMSE} & \textbf{sMAPE} & \textbf{DM test DNN Enemble} & \textbf{DM test DNN Enesemble renorm last year} & \textbf{DM test Naive Model} \\
              \midrule
              Nord Pool & 2.33 & 4.08 & 0.07 & 0 & 0 & 0 \\ 
              PJM & 3.67 & 5.85 & 0.14 & 0 & 0 & 0  \\ 
              EPEX-BE & 6.54 & 16.68 & 0.15 & 0 & 0 & 0  \\ 
              EPEX-FR & 4.91 & 12.67 & 0.14 & 0.97 & 1 & 0  \\ 
              EXPEX-DE & 4.03 & 6.99 & 0.17 & 0 & 0 & 0  \\
            \bottomrule
        \end{tabular}
    }
\end{table}

\begin{figure*}
  \centering
  \begin{subfigure}{0.3\linewidth}
      \centering
      \includegraphics[width=\linewidth, height=0.8\linewidth]{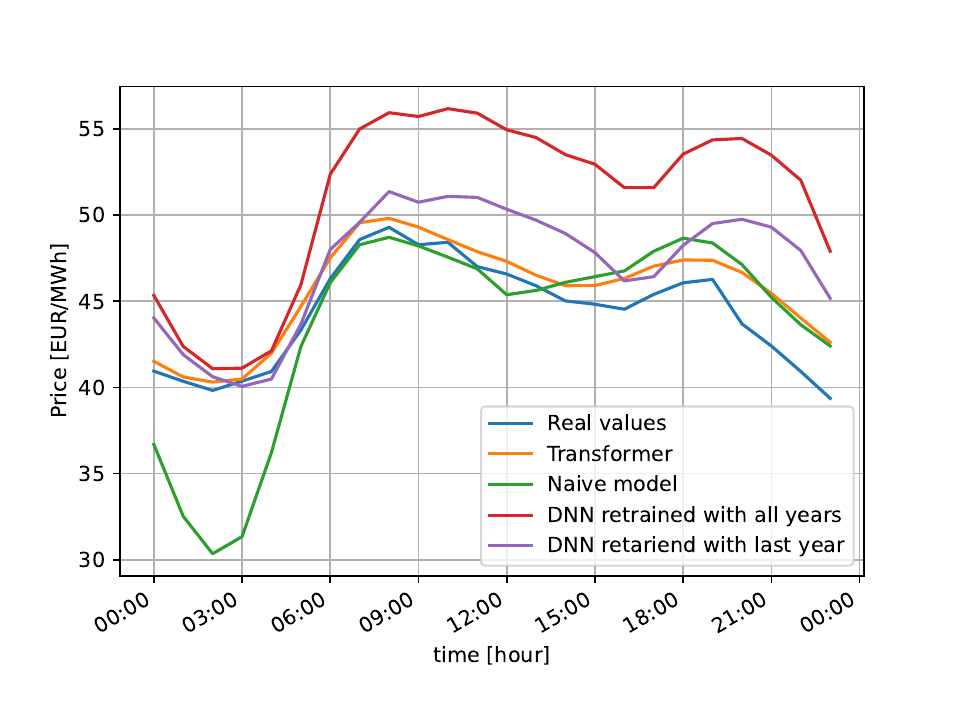}
      \caption{Nord Pool}
  \end{subfigure}
  \begin{subfigure}{0.3\linewidth}
      \centering
      \includegraphics[width=\linewidth, height=0.8\linewidth]{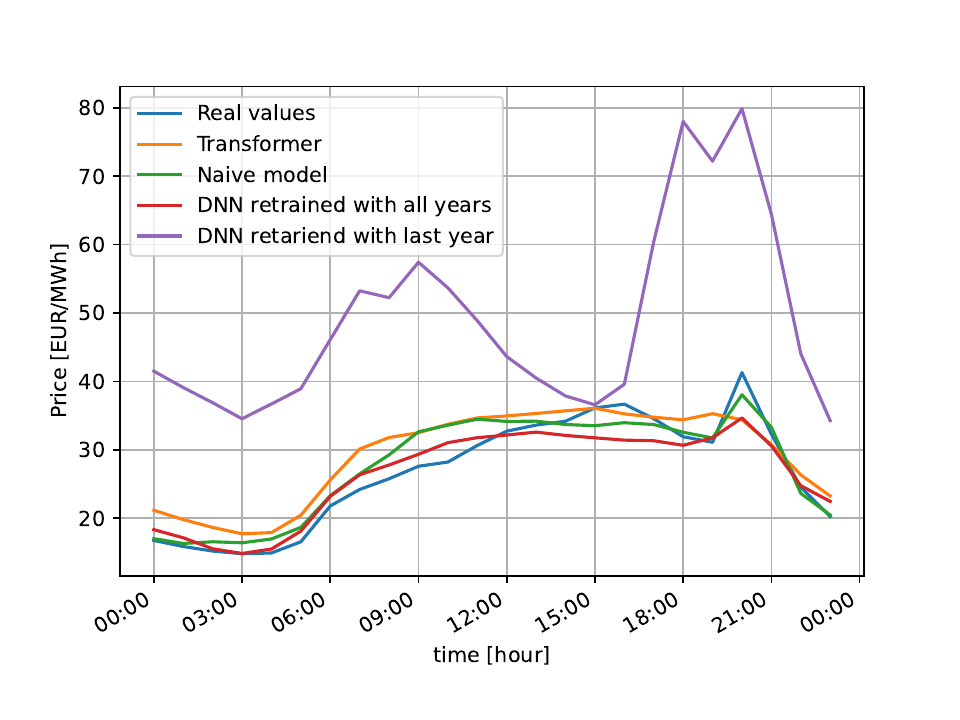}
      \caption{PJM}
  \end{subfigure}
  \begin{subfigure}{0.3\linewidth}
      \centering
      \includegraphics[width=\linewidth, height=0.8\linewidth]{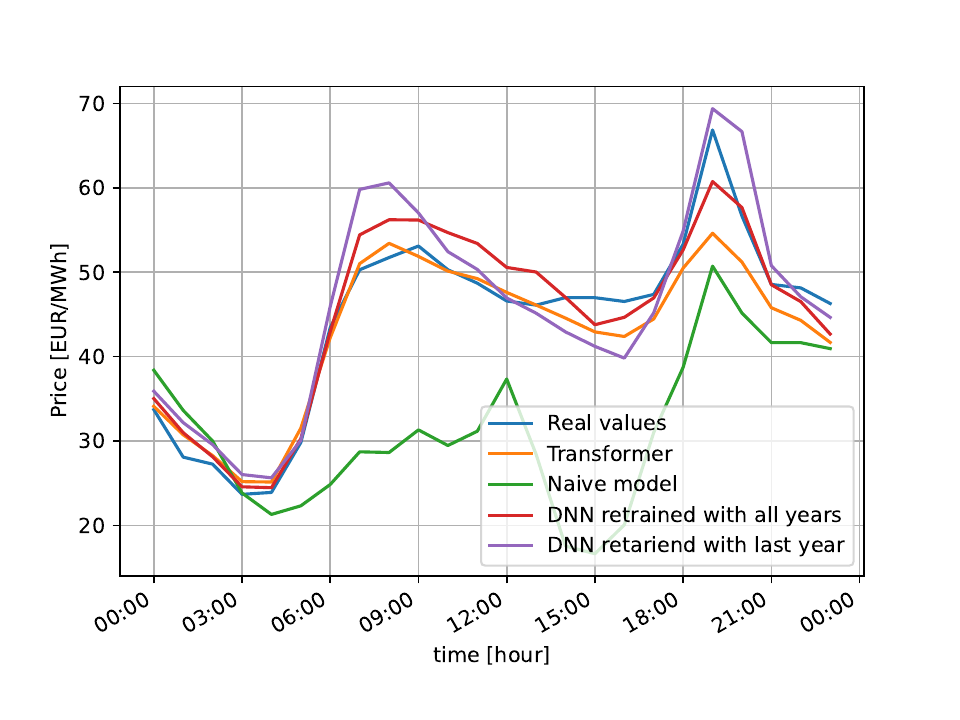}
      \caption{EPEX-FR}
  \end{subfigure}
  \begin{subfigure}{0.3\linewidth}
      \centering
      \includegraphics[width=\linewidth, height=0.8\linewidth]{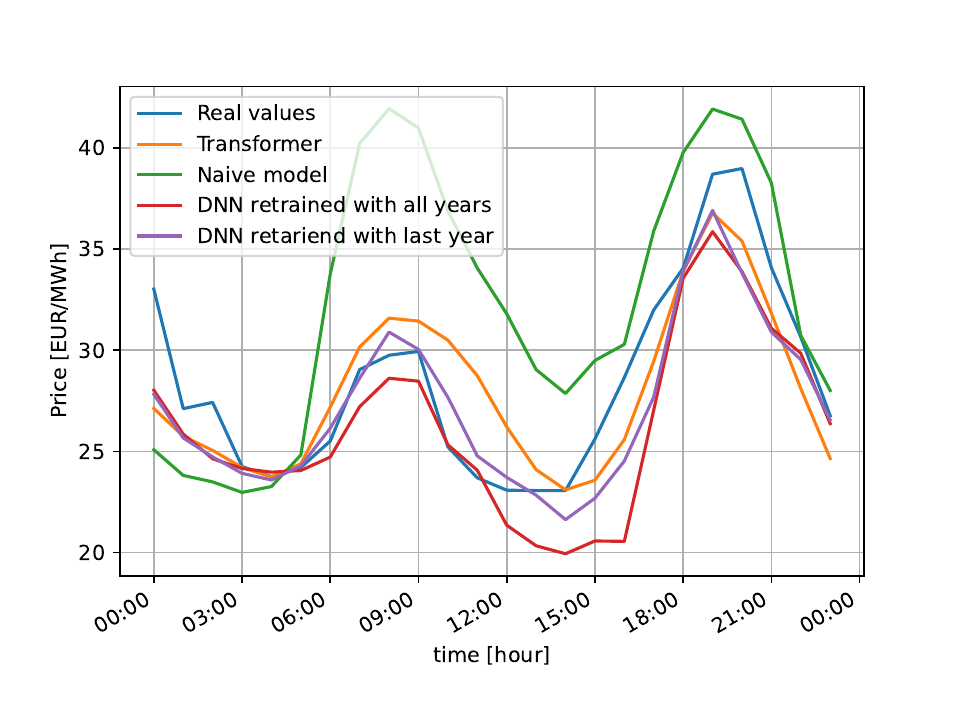}
      \caption{EPEX-DE}
  \end{subfigure}
  \begin{subfigure}{0.3\linewidth}
      \centering
      \includegraphics[width=\linewidth, height=0.8\linewidth]{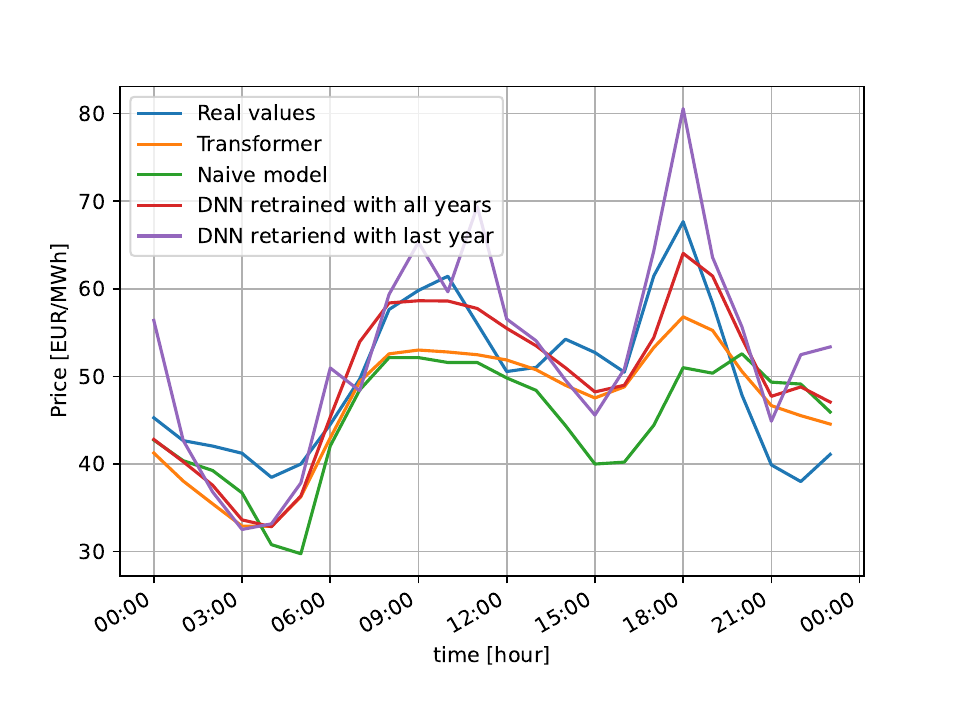}
      \caption{EPEX-BE}
  \end{subfigure}
  \caption{Example predictions for one day}
  \label{fig: example predictions}
\end{figure*}

An example of a prediction for one day per dataset can be visualized in Figure~\ref{fig: example predictions}. It can be observed that even if in some cases it is not the best prediction, our model is more consistent than the other models, predicting an accurate forecast in every example. This is in line with the results of the metrics explained above, being the more consistent model the one that have the best results on average (MAE, RMSE and sMAPE).

\section{Conclusion and Future work}
Even though the Transformer has been used for forecasting in other domains, there was a lack of conclusive evidence that indicated the performance against other approaches for electricity price forecasting. In this paper, a transformer-based forecasting model is proposed and tested in the framework proposed in \cite{lagoForecastingDayaheadElectricity2021}. It has been shown that the model offers significant better performance for the majority of the cases, having state-of-the-art results for four of the five datasets. Hence, the Transformer architecture offers good prediction results for electricity price forecasting. 

Regarding future work, there could be several lines of research worth exploring. For example, the authors in \cite{lagoForecastingDayaheadElectricity2021} claim that LSTMs could potentially be more accurate than DNN, but the evidence was insufficient. A possible future study would be a comparison between LSTMs and Transformers. It is important to note that the model presented in this paper is not an Ensemble. An Ensemble of Transformers could have been better, but it was out of the scope of this paper. However, it would be a relevant field of study. Finally, another future work would be to explore how to take also into account past values of the exogenous variables in the transformer arquitecture.
 





\bibliographystyle{cas-model2-names}

\bibliography{main}

\appendix

\section{Validation results for the proposed model arquitecture}

\begin{table}
    \centering
    \caption{Nord Pool Validation Results}
    \label{tab: nord pool validation set results}
    \begin{tabular}{ccccccccccc}
        \toprule
          \textbf{$n$} & \textbf{NH} & \textbf{NL} & \textbf{FD} & \textbf{P} & \textbf{Opt} & \textbf{SL} & \textbf{Lr} & \textbf{Sch} & \textbf{CG} & \textbf{MAE}\\ 
          \midrule
          128 & 1 & 6 & 1024 & 0.2 & Adam & 672 & 1e-4 & 70-0.1 & None & 1.336 \\ 
          512 & 8 & 6 & 2048 & 0.2 & Adam & 672 & 1e-5 & 70-0.1 & None & 1.264 \\ 
          \rowcolor{lightgray} 512 & 8 & 6 & 1024 & 0.2 & Adam & 336 & 1e-5 & 70-0.1 & None & 1.196 \\ 
          256 & 8 & 6 & 1024 & 0.1 & Adam & 672 & 1e-5 & 70-0.1 & None & 1.225 \\ 
          128 & 8 & 6 & 512 & 0.1 & Adam & 672 & 1e-5 & 70-0.1 & None & 1.243 \\ 
        \bottomrule
    \end{tabular}
\end{table}

\begin{table}[h]
    \centering
      \caption{PJM Validation Results}
      \label{tab: pjm validation set results}
    \begin{tabular}{ccccccccccc}
        \toprule
          \textbf{$n$} & \textbf{NH} & \textbf{NL} & \textbf{FD} & \textbf{P} & \textbf{Opt} & \textbf{SL} & \textbf{Lr} & \textbf{Sch} & \textbf{CG} & \textbf{MAE}\\ 
          \midrule
          \rowcolor{lightgray} 128 & 4 & 4 & 1048 & 0.2 & Adam & 336 & 1e-4 & 70-0.1 & 0.1 & 2.526 \\ 
          64 & 2 & 4 & 256 & 0.2 & Adam & 336 & 1e-4 & 70-0.1 & 0.1 & 2.576 \\ 
          64 & 4 & 4 & 1024 & 0.2 & Adam & 336 & 1e-4 & None & 0.1 & 2.626 \\ 
          64 & 4 & 4 & 1024 & 0.2 & Adam & 336 & 1e-3 & 70-0.1 & 0.1 & 2.699 \\ 
          64 & 8 & 6 & 1024 & 0.2 & Adam & 336 & 1e-5 & None & 0.1 & 2.811 \\ 
        \bottomrule
    \end{tabular}
\end{table}

\begin{table}
    \centering
      \caption{EPEX-FR Validation Results}
      \label{tab: epex-fr validation set results}
    \begin{tabular}{ccccccccccc}
        \toprule
          \textbf{$n$} & \textbf{NH} & \textbf{NL} & \textbf{FD} & \textbf{P} & \textbf{Opt} & \textbf{SL} & \textbf{Lr} & \textbf{Sch} & \textbf{CG} & \textbf{MAE}\\ 
          \midrule
          512 & 1 & 6 & 1024 & 0.2 & Adam & 672 & 1e-5 & 50-0.2 & 0.25 & 3.778 \\ 
          512 & 1 & 6 & 2048 & 0.2 & Adam & 672 & 1e-5 & 50-0.2 & 0.25 & 3.853 \\ 
          512 & 8 & 6 & 1024 & 0.2 & Adam & 336 & 1e-5 & 50-0.2 & 0.25 & 3.756 \\ 
          \rowcolor{lightgray} 512 & 8 & 6 & 1024 & 0.2 & Adam & 672 & 1e-5 & 50-0.2 & 0.25 & 3.732 \\ 
          512 & 8 & 6 & 1024 & 0.2 & Adam & 672 & 1e-5 & 70-0.2 & 0.25 & 3.751 \\ 
          512 & 8 & 6 & 102 & 0.2 & Adam & 672 & 1e-5 & 70-0.1 & 0.5 & 3.759 \\ 
        \bottomrule
    \end{tabular}
\end{table}

\begin{table}[h]
    \centering
      \caption{EPEX-DE Validation Results}
      \label{tab: epex-de validation set results}
    \begin{tabular}{ccccccccccc}
        \toprule
          \textbf{$n$} & \textbf{NH} & \textbf{NL} & \textbf{FD} & \textbf{P} & \textbf{Opt} & \textbf{SL} & \textbf{Lr} & \textbf{Sch} & \textbf{CG} & \textbf{MAE}\\ 
          \midrule
          256 & 4 & 4 & 1024 & 0.2 & Adam & 336 & 1e-4 & None & 0.5 & 3.904 \\ 
          \rowcolor{lightgray} 256 & 8 & 6 & 1024 & 0.2 & Adam & 336 & 1e-5 & None & 0.25 & 3.577 \\ 
          512 & 8 & 6 & 2048 & 0.2 & Adam & 672 & 1e-5 & None & 0.25 & 3.886 \\ 
          512 & 8 & 6 & 2048 & 0.2 & Adam & 672 & 1e-6 & None & 0.25 & 3.821 \\ 
          64 & 4 & 4 & 256 & 0.2 & Adam & 336 & 1e-5 & None & 0.5 & 3.887 \\ 
          64 & 8 & 6 & 256 & 0.2 & Adam & 336 & 1e-5 & None & 0.25 & 3.873 \\ 
        \bottomrule
    \end{tabular}
\end{table}

\begin{table}[h]
    \centering
      \caption{EPEX-BE Validation Results}
      \label{tab: epex-be validation set results}
    \begin{tabular}{ccccccccccc}
        \toprule
          \textbf{$n$} & \textbf{NH} & \textbf{NL} & \textbf{FD} & \textbf{P} & \textbf{Opt} & \textbf{SL} & \textbf{Lr} & \textbf{Sch} & \textbf{CG} & \textbf{MAE}\\ 
          \midrule
          128 & 4 & 4 & 1024 & 0.2 & Adam & 672 & 1e-5 & None & 0.25 & 5.238 \\ 
          128 & 8 & 6 & 1024 & 0.2 & Adam & 672 & 1e-5 & None & 0.25 & 5.265 \\ 
          64 & 4 & 4 & 1024 & 0.2 & Adam & 672 & 1e-5 & None & 0.5 & 5.324 \\ 
          64 & 4 & 4 & 256 & 0.2 & Adam & 336 & 1e-4 & None & 0.1 & 5.125 \\ 
          64 & 4 & 4 & 256 & 0.2 & Adam & 336 & 1e-5 & None & 0.1 & 4.903 \\ 
          \rowcolor{lightgray} 64 & 4 & 4 & 256 & 0.2 & Adam & 672 & 1e-5 & None & 0.1 & 4.854 \\ 
        \bottomrule
    \end{tabular}
\end{table}

\end{document}